\documentclass[conference]{IEEEtran}
\IEEEoverridecommandlockouts

\usepackage{amsmath,amssymb,amsfonts}
\usepackage{graphicx}
\usepackage{microtype}
\usepackage{subfigure}
\usepackage{hyperref}
\usepackage{booktabs} 
\usepackage{multirow}
\usepackage{makecell}
\usepackage{algorithm}
\usepackage[noend]{algpseudocode}

\usepackage{mathtools}
\usepackage{amsthm}

\DeclareMathOperator\erf{erf}

\begin{document}

\title{A Data-Free Analytical Quantization Scheme for Deep Learning Models\thanks{This paper has been accepted at the IEEE International Conference on Data Mining (ICDM) 2025.}}

\author{\IEEEauthorblockN{Ahmed Luqman}

\and
\IEEEauthorblockN{Khuzemah Qazi}

\and
\IEEEauthorblockN{Murray Patterson}

\and

\IEEEauthorblockN{Malik Jahan Khan}

\and
\IEEEauthorblockN{Imdadullah Khan}

}

\author{
    \IEEEauthorblockN{Ahmed Luqman\IEEEauthorrefmark{1},
    Khuzemah Qazi\IEEEauthorrefmark{1},
    Murray Patterson\IEEEauthorrefmark{2},
    Malik Jahan Khan\IEEEauthorrefmark{1},
    Imdadullah Khan\IEEEauthorrefmark{1}}
    
    \IEEEauthorblockA{\IEEEauthorrefmark{1}Department of Computer Science, Lahore University of Management Sciences, Lahore, Pakistan \\Email: \{24100041, 24100092, jahan, imdad.khan\}@lums.edu.pk}
    
    
    \IEEEauthorblockA{\IEEEauthorrefmark{2}Department of Computer Science, Georgia State University, Atlanta, 30303, GA, USA \\ Email: mpatterson30@gsu.edu}


}

\maketitle

\begin{abstract}
Despite the success of CNN models on a variety of Image classification and segmentation tasks, their extensive computational and storage demands pose considerable challenges for real-world deployment on resource-constrained devices. Quantization is one technique that aims to alleviate these large storage requirements and speed up the inference process by reducing the precision of model parameters to lower-bit representations. In this paper, we introduce a novel post-training quantization method for model weights. Our method finds optimal clipping thresholds and scaling factors along with mathematical guarantees that our method minimizes quantization noise. Empirical results on real-world datasets demonstrate that our quantization scheme significantly reduces model size and computational requirements while preserving model accuracy.
\end{abstract}


\section{Introduction}

	Deep neural networks (DNNs) are fundamental to modern artificial intelligence, driving advancements in fields like computer vision, natural language processing, and autonomous systems. However, the deployment of DNNs in real-world applications is often limited by their significant computational and storage demands, primarily due to the vast number of model parameters. This is especially true for Convolutional Neural Networks (CNNs) used in image classification and segmentation, which have numerous layers, leading to substantial storage requirements. These demands make it challenging to deploy full models on devices with limited resources.
		
    A practical solution to address the growing complexity of DNNs is model compression, which reduces the size and computational demands of these networks with minimal impact on performance. By compressing a model, we create a smaller, less complex version that maintains key performance metrics like accuracy and loss. The advantages of model compression include reduced model size, lower computational complexity, decreased memory usage, improved energy efficiency, and faster inference times. This enables efficient deployment of DNNs on resource-constrained devices and can, in some cases, improve generalization by mitigating overfitting.

    Model compression can be achieved through techniques such as pruning (removing redundant parameters), knowledge distillation (transferring knowledge from a larger model to a smaller one), and matrix factorization (decomposing weight matrices into low-rank approximations). In our paper, we focus on quantization, a method that reduces the precision of numerical values in model parameters. Quantization can be applied to both the weights and activations of a DNN. It involves converting high-precision weights and biases (e.g., 32-bit floating point) to lower precision (e.g., 8-bit integers), and quantizing intermediate activations during inference. This enables low-precision computation, leveraging fixed-bit arithmetic for efficient hardware implementation, making it feasible to deploy models on edge devices. Recent research has explored enhancing quantization by combining it with other compression techniques, further improving its effectiveness~\cite{ko2021mascot, wang2020differentiable}.

	Quantization Aware Training (QAT) incorporates quantization directly into the training process, allowing the model to ``learn'' the effects of quantization. However, this is time-consuming and difficult to implement~\cite{gholami2022survey}. Post-Training Quantization (PTQ) adjusts the weights of the quantized model without the need for further training~\cite{banner2019post}. PTQ can be applied to both weights and activations. 
		
    Uniform quantization divides the parameter value range into equal-sized intervals, offering a simple and efficient method. However, it may be suboptimal for parameters with non-uniform distributions. Non-uniform quantization, in contrast, adjusts interval sizes based on the parameter distribution, potentially offering a better fit for data with long tails, which is common for weights and activations. This approach can improve accuracy by assigning more precision to frequently occurring values. However, it is more complex, may lack compatibility with certain hardware, and could increase computational costs, limiting its portability and efficiency.
		
		In this paper, we propose a novel post-training quantization approach that uniformly quantizes model weights and biases to any user-defined number of bits. Since model parameters generally have bell-shaped distribution, most commonly Laplace or Gaussian distribution~\cite{baskin2021uniq,li2019additive,banner2019post}, we derive an analytical solution for uniform quantization. We determine the optimal clipping range, quantization intervals, and quantization levels, ensuring minimal quantization error.
		
		
		The key contributions of this work are as follows:
		\begin{itemize}
			\item Demonstrating that weights and activations predominantly follow either a Gaussian or Laplacian distribution.
			\item Introducing a numerical quantization solution for non-uniform distributions with analytical proof.
			\item Showing that our quantization scheme outperforms existing methods in terms of mean square error (MSE) performance.
			\item Providing evidence that well-known models quantized using our scheme exhibit minimal loss of accuracy.
		\end{itemize}
		
		The remainder of this paper is organized as follows: Section~\ref{sec_Related} reviews related work. Section~\ref{sec_proposed} elaborates on the details of our method. Section~\ref{sec:experiments} presents the experimental results and analysis. Finally, Section~\ref{sec:conclusion} discusses the conclusion and future work.

		
		


		\section{Related Work}\label{sec_Related}
  
  This section outlines some related work on model compression with a focus on quantization. The challenge of reducing model size, improving inference latency, and minimizing power consumption in deep neural networks is addressed \cite{gholami2022survey} through knowledge distillation methods \cite{gou2021knowledge} and matrix factorization \cite{ko2021mascot,Gao2021_ICDM_Dict_DFQ, Gao2022ICDM_DFQ}, and pruning \cite{chen2024optimal}.

		\cite{cho2024pdp} introduces an efficient train-time pruning scheme that uses a dynamic function of weights to generate soft pruning masks without additional parameters. In \cite{el2022data}, the authors introduce a data-efficient structured pruning method based on submodular optimization. This ensures minimal input changes in the next layer and provides an exponentially decreasing error with pruning size. Hardware-aware latency Pruning \cite{shen2022structural} formulates structural pruning as a global resource allocation optimization problem, aiming to maximize accuracy while keeping latency within a predefined budget on the target device. 
		
		Network quantization compresses and accelerates networks by mapping floating-point weights to lower-bit integers. Activations are only known when the network is run on training data; activation quantization can only be done using calibration data. Selection of calibration training data is critical.  Williams et al. \cite{williams2023does} demonstrate how random sampling to get calibration data is not optimal. Chen et al. \cite{chen2022climbq} shows the need to prevent class-imbalanced calibration data. Zhang et al. \cite{zhang2023selectq} propose a method to get a truly representative sample to be used to get correct statistics. 
		
		Quantization Aware Training (QAT) incorporates quantization effects during training, allowing models to adapt to lower precision \cite{ahmadian2024intriguing}. Post Training Quantization (PTQ), on the other hand, adjusts quantized model weights post-training, eliminating the need for additional training cycles and access to training data.

Data-Free Quantization (DFQ) extends PTQ by generating synthetic data to calibrate the quantized model when real training data is unavailable. Traditional DFQ approaches typically optimize a generator to mimic the full-precision model, without explicitly accounting for how informative the generated samples are for the quantized model. To address this limitation, Qian et al. \cite{qian2023adaptive} introduce Adaptive Data-Free Quantization (AdaDFQ), which frames the generation of calibration data as a zero-sum game between a generator and a quantized network. By optimizing the margin between ``agreement" and ``disagreement" samples, AdaDFQ adaptively controls the informativeness of generated samples to avoid overfitting and improve generalization. Their findings challenge the assumption that maximum adaptability always leads to better quantized performance and emphasize the importance of aligning synthetic sample informativeness with category and distribution characteristics of the original dataset.
		 
		\cite{sun2022entropy} proposed entropy-driven mixed-precision quantization, which assigns optimal bit-widths to each layer based on weight entropy, facilitating deployment on IoT devices. In \cite{tukan2022pruning}, the authors propose a robust, data-independent framework for computing coresets with mild model assumptions, using Löwner ellipsoid and Caratheodory theorem to assess neuron importance. The method works across various networks and datasets, achieving a $62\%$ compression rate on ResNet50 with only a $1.09\%$ accuracy drop on ImageNet. In \cite{ma2023ompq},  mixed precision quantization is proposed, which utilizes hardware’s multiple bit-width arithmetic operations. Kuzmin et al. \cite{kuzmin2022fp8} showed how having 8 bits does not necessitate using integers and that FP8 can perform better than INT8.
		
		Uniform quantization, a straightforward approach, involves partitioning the range of parameter values into equal-sized intervals \cite{banner2019post}. Neural Network distributions tend to have a bell-curved distribution around the mean \cite{baskin2021uniq}. Several papers take advantage of this and allocate bins accordingly. Banner et al. \cite{banner2019post} introduce Analytical Clipping for Integer Quantization (ACIQ), which limits activation value ranges to reduce rounding error by minimizing mean-square-error. They also propose a bit allocation policy to optimize bit-width for each channel, formulated as an optimization problem to minimize overall mean-squared error under quantization constraints. Their approach achieves accuracy that is just a few percent less than the state-of-the-art baseline across a wide range of convolutional models. Zhao et al. \cite{pmlr-v97-zhao19c} expand on \cite{banner2019post} by artificially shifting outliers closer to the center of the distribution.
		
		Nonuniform quantization, which adapts interval sizes to the parameter distribution, generally outperforms uniform quantization in compressing neural networks due to its better representational capacity ~\cite{liu2022nonuniform}. Nonuniform quantization allocates more precision to frequently occurring values, enhancing accuracy for non-uniform data. However, the resulting model's complexity can be increased, and hardware compatibility issues can make it more complex. 
  
  ~\cite{li2019additive} proposed a method to discretize parameters in a non-uniform manner to additive powers of two, aligning more closely with their distribution and improving efficiency.

		Our proposed method contributes to this work by providing an analytical framework for non-uniform quantization that does not depend on training data. This independence is particularly advantageous for scenarios where data availability is limited or privacy concerns restrict the use of actual data for model compression.

		\section{Proposed Method}\label{sec_proposed} 
        Our method first uses the Kolmogorov-Smirnov test to evaluate whether the given model has weights that more closely resemble a Normal Distribution or a Gaussian Distribution. After finding the best fitting distribution, we use it to calculate the optimal quantization levels and intervals for each layer of the network.
		Finding these intervals is computationally intensive as it involves solving a system of linear equations, so we use an iterative approach instead. 


		\subsection{Preliminaries}
		Let $X$ be a high-precision tensor-valued random variable with a probability density function $f(x)$. Assuming target bit-width $M$, we clip the range of values in the tensor and quantize the values in the clipped range to $2^M$ discrete values. We give an analytical expression for expected clipping and quantization error. We find the optimal clipping range, quantization intervals, and quantization level for each interval that minimize the sum of clipping and quantization errors for the cases when $f(x)$ is a Gaussian or Laplace distribution. 
		
		We formulate the quantization problem as an optimization task with the objective of minimizing Mean Squared Error (MSE).
		
		\subsubsection{Clipping Error}
  
		For $x \in \mathbb{R}$ and constants $\alpha < \beta$,  define the clip function 
  
\[
\text{clip}(x,\alpha,\beta) = \begin{cases} 
\alpha & \text{if } x < \alpha, \\
x & \text{if } \alpha \leq x \leq \beta, \\
\beta & \text{if } x > \beta.
\end{cases}
\]

 $\alpha$, and all weights above $\beta$ would become $\beta$. 
		
		The clipping error $C_E$, the mean-squared error resulting from clipping, is given as:
		\begin{align}
			C_E &= C_E(\alpha,\beta) =\int_{-\infty}^{\infty} \big(x - clip(x,\alpha,\beta)\big)^2 f(x) \, dx \notag \\[.03in]
			=&\int_{-\infty}^{\alpha} (x - \alpha)^2 f(x) \, dx + \int_{\beta}^{+\infty} (x - \beta)^2 f(x) \, dx
		\end{align}
		
		Using the knowledge of the weight distribution, we propose a method to calculate the quantization intervals $(x_i$) and levels $(y_i)$ that minimize the MSE. We introduce a clipping range with limits $\alpha$ and $\beta$ to handle weights that fall outside the typical distribution range, reducing the clipping error.

		\subsubsection{Quantization Error}
		Let $X'$ be the clipped tensor, i.e., $X$ clipped to the range $[\alpha,\beta]$. The range $[\alpha,\beta]$ is partitioned into $k = 2^M$ intervals $[x_0,x_1),[x_1,x_2],\ldots,(x_{k-1},x_k]$, where $x_0 = \alpha$ and $x_k = \beta$. For $i=1,\cdots,k$, all the values in $X'$ that fall in $[x_{i-1},x_i)$ are rounded to $y_i \in [x_{i-1},x_i)$. The quantization error in the range  $[\alpha,\beta]$ is given as 
		\begin{equation}
			Q_E =\sum_{i=1}^{2^M} \int_{x_{i-1}}^{x_i} (x - y_i)^2 f(x) \, dx
		\end{equation}

		\subsection{Optimization Framework}
		Our quantization method involves determining optimal clipping thresholds and scaling factors that minimize quantization noise. Our goal is to find the optimal clipping range ($[\alpha,\beta]$), quantization intervals (given by $x_1,\ldots,x_{k-1}$), and quantization levels (given by $y_1,\ldots,y_k$). The expected mean square error between $X$ and its clipped and quantized version $Q(X)$ is given as follows:
		
		\begin{align}\label{clip_Quantization_Error}
			&D = E[(X - Q(x))^2] \notag{}\\[.05in]
			& = \int_{-\infty}^{\alpha} (x - \alpha)^2 f(x) \, dx + \int_{\beta}^{+\infty} (x - \beta)^2 f(x) \, dx \notag{}\\[.05in] 
			& + \sum_{i=1}^{2^M} \int_{x_{i-1}}^{x_i} (x - y_i)^2 f(x) \, dx
		\end{align}

		The mathematical model for non-uniform quantization is based on the probability distribution of the data. Given a set of data points $X$  with a probability density function $f(x)$, the goal is to find a quantization function $Q(x)$ that minimizes the expected distortion $D$, defined as:
		
		\begin{equation}
			D = E[(X - Q(X))^2]
		\end{equation} 
		
		The quantization function $Q(x)$ maps the data points to a set of quantization levels $\{y_i\}$, with the intervals $\{I_i=[x_{i-1},x_i]\}$ determined by the distribution $f(x)$. The optimization problem can be formulated as:
		\begin{equation}\label{objFunction}
			\arg\min_{\{y_i\}, \{I_i\}} E[(X - Q(X))^2] 
		\end{equation}

		\subsubsection{Optimal Quantization Levels for Known Quantization Intervals}
		
		Assuming that quantization intervals ($x_i, i = 1,2,...,K$ ) are known, we need to determine the quantization levels ($y_i, i = 1,2,...,K$). 
		
		We minimize the $D$ \eqref{clip_Quantization_Error} with respect to $y_i$. Since $D$ is a sum of integrals, and only one of those integrals includes $y_i$ (the one with limits $x_{i-1}$ and $x_i$), the other terms differentiate to 0.

		\begin{equation}
			\frac{\delta Q_E}{\delta y_i}  = -2 \int_{x_{i-1}}^{x_i} (x-y_i)f(x) \, dx = 0
		\end{equation}  
		
		We open up this integral and evaluate this in terms of $y_i$:
		
		\begin{align}
			&\int_{x_{i-1}}^{x_i} xf(x) \, dx - \int_{x_{i-1}}^{x_i} y_i f(x) \, dx  = 0  \notag{}\\[.05in]
			\implies &\int_{x_{i-1}}^{x_i} xf(x) \, dx = y_i \int_{x_{i-1}}^{x_i} f(x) \, dx    \notag{}\\[.05in]
			\implies &y_i = \frac{\int_{x_{i-1}}^{x_i} xf(x) \, dx }{\int_{x_{i-1}}^{x_i} f(x) \, dx }
		\end{align}  
		
		In other words, we get that $y_i$ should be the conditional expected value of the weights within the interval $[x_{i-1},x_i)$.
		
		\begin{equation}
			y_i =  E[X|x_{i-1} < X < x_i]
			\label{conditional_mean_criterion}
		\end{equation}  
		
		We call this our Conditional Mean Criterion. This means that all values that fall in a certain interval should be quantized to their expected value in that interval. Many quantization methods like ACIQ \cite{banner2019post} resort to simply quantizing the values to the midpoint of the relevant interval, which is only correct should the distribution be uniform.

		\subsubsection{Optimal Quantization Intervals for Known Quantization Levels}
		
		Assuming that quantization levels  ($y_i, i = 1,2,...,K$ ) are known, we need to know how to allocate the quantization intervals ($x_i, i = 1,2,...,K$). We minimize the $D$  \eqref{clip_Quantization_Error} with respect to $x_i$.
		
		Since $D$ is a sum of integrals, and only two of those integrals include $x_i$ as a limit, the rest is differentiated to 0.
		\resizebox{.99\linewidth}{!}{
			\begin{minipage}{\linewidth}
				\begin{align} 
					&\frac{\delta Q_E}{\delta x_i} \notag{}\\
					&=\frac{\delta}{\delta x_i} \int_{x_{i-1}}^{x_i} (x - y_i)^2 f(x) \, dx + \frac{\delta}{\delta x_i} \int_{x_{i}}^{x_{i+1}} (x - y_{i+1})^2 f(x) \, dx \notag{} \\
					&= 0
					\label{unsimplified_midpoint}
				\end{align} 
			\end{minipage}
		}

		Since we are differentiating an integral with respect to its limits, we use the Leibniz Rule \eqref{eq:leibniz_basic}:
		
		\begin{equation}
			\frac{d}{dx} \int_a^x f(t) \, dt = f(x) \cdot \frac{dx}{dx} = f(x)
			\label{eq:leibniz_basic}
		\end{equation} 
		
		Using the Leibniz Rule  \eqref{eq:leibniz_basic} to simplify \ref{unsimplified_midpoint}:
		
		\begin{equation}
			\frac{\delta Q_E}{\delta x_i}  = (x_i - y_i)^2f(x_i) - (x_i - y_{i+1})^2f(x_i) = 0
		\end{equation}

		Evaluating this in terms of $x_i$ gives us our Midpoint Criterion. This shows us that the boundary should be the midpoint of the quantization levels, regardless of what $f(x)$ is. 
		
		\begin{equation}
			(x_i - y_i)^2 =  (x_i - y_{i+1})^2
		\end{equation}  
		
		\begin{equation}
			x_i = \frac{y_i + y_{i+1}}{2}, i = 1,2,...,K-1
			\label{midpoint_criterion}
		\end{equation}  
		
		A simple calculation shows that regardless of the distribution of $X$, the above quantization interval's boundaries minimize the $D$ for the given $y_i$. This seems to contradict the Conditional Mean Criterion; however, we can get a system of equations using these two criteria.
		
		\subsubsection{System of Simultaneous Equations}

		Given the general forms of the two criteria apply for all values of $i = 1,2,...,K$, we get a system of $2K-1$ equations. These can be solved to get the $2K-1$ unknowns, which are the quantization levels and boundaries.
		
		Solving this large system of equations has high computational complexity; hence, we choose to compute the optimum quantizer in an iterative fashion. We start by arbitrarily setting the quantization bins in a uniform fashion. We choose to start with $x_i = 2i/M$. We then iterate between computing new quantization levels according to Equation 8 and new quantization bin edges according to Equation 13. After going back and forth between these two equations several times, the results converge toward a final optimum quantizer.

  \subsubsection{Iterative approach to solving the system}

  	Due to the high computational cost of solving a large system of equations directly, we employ an iterative approach to find the optimal quantizer. \ref{alg:quantizer_opt}. 
   
		We begin by initializing the quantization bins uniformly across the range. A common starting point is to set each bin center, denoted by $x_i$, to $2i/K$, where $i$ ranges from 1 to $K$ (the number of bins). The optimization process then iterates between two steps:
		
		1. Update Quantization Levels: We calculate new, improved quantization levels based on Equation \ref{conditional_mean_criterion}.
		
		2. Update Bin Edges: We compute new and refined bin boundaries using Equation \ref{midpoint_criterion}.
		
		By repeatedly going back and forth between these steps, the initial estimate progressively converges towards the optimal quantizer configuration.
		
\begin{algorithm}
\caption{Iterative Quantizer}
\label{alg:quantizer_opt}
\begin{algorithmic}[1]
    \State {\bfseries Input:} Data points $\{x_n\}$, number of bins $K$
    \State {\bfseries Initialize:} Quantization levels $y_i = \frac{2i}{K}$ for $i = 1, \dots, K$  
    
    \While{Stopping criterion not met}
        \State {\bfseries Update Quantization Levels:}
        \For{$i = 1$ {\bfseries to} $K$}
            \State $y_i \leftarrow$ UpdateLevel($y_i$)
            \Comment{Using Eq.~\ref{conditional_mean_criterion}}
        \EndFor
        \State {\bfseries Update Quantization Boundaries:}
        \For{$i = 1$ {\bfseries to} $K - 1$}
            \State $x_i = \frac{y_i + y_{i+1}}{2}$
            \Comment{Using Eq.~\ref{midpoint_criterion}}
        \EndFor
        \State $x_K = \infty$  
    \EndWhile
    
    \State {\bfseries Output:} Optimal quantization levels $\{y_i\}$ and boundaries $\{x_i\}$
\end{algorithmic}
\end{algorithm}

\subsection{Convergence of the Iterative Algorithm}
	As we iteratively update the quantization levels and boundaries, we obtain increasingly smaller intervals at the center, resulting in an increasing number of intervals in close to the center.

Convergence relies on the monotonic decrease in MSE that follows error minimization [Equation (6)]. We perform two steps in each iteration, updating (1) quantization levels and (2) intervals using conditional mean and midpoint criteria, respectively. For fixed intervals, the conditional expectation minimizes MSE in each interval. For fixed quantization levels, setting the interval boundaries at the midpoints minimizes the MSE between quantization levels.

Since the MSE is bounded below and decreases in each iteration, the sequence converges to a limit. In line with this, experiments on all models and datasets showed a strictly decreasing MSE with each successive iteration.

		

		\subsection{Analytical Solutions for Bell-Shaped Distributions}
		We consider two common bell-shaped distributions: Gaussian and Laplace \cite{banner2019post}. The intervals $\{I_i=[x_{i-1}, x_i)\}$ are chosen to reflect the concentration of data points, with more intervals placed where the data is denser.
		\subsubsection{Laplace Distribution}
		The Laplace distribution is particularly well-suited for modeling data that exhibit a sharp peak at the mean, with heavy tails decaying exponentially on both sides. This characteristic makes it a compelling choice for quantizing neural network parameters, which often follow a similar distribution.
		
		The Laplace distribution is defined by its location parameter $\mu$, which denotes the peak of the distribution, and its diversity parameter $b$, which controls the decay rate of the tails. The probability density function (PDF) of the Laplace distribution is given by:
  
		\begin{equation}
			f(x|\mu, b) = \frac{1}{2b} \exp\left(-\frac{|x - \mu|}{b}\right)
		\end{equation}  
		The sharp peak at the mean and the heavy tails are advantageous for quantization as they allow for a more concentrated set of intervals around the mean, where data points are most dense, and larger intervals in the tails, where data points are sparse.
		
		To derive the analytical solution for quantization under the Laplace distribution, we seek to minimize the expected value of the quantization error, which can be expressed as:
		\begin{equation}
			E[Q_E] = \int_{-\infty}^{\infty} (x - Q(x))^2 f(x|\mu, b) dx
		\end{equation}  
		where $Q(x)$ is the quantization function mapping real numbers to a discrete set of quantization levels. By leveraging the symmetry and the exponential decay properties of the Laplace distribution, we can determine the optimal quantization levels and intervals that minimize this error. The resulting quantization scheme is non-uniform, with smaller intervals near the mean and progressively larger intervals as we move away from the mean, reflecting the density of the distribution.
		The quantization level $y_i$ with the standard Laplacian parameters is as follows:
		\begin{align} \label{eq:laplace_opt_ppoints}
			y_i &= \frac{\int_{x_{i-1}}^{x_i} xe^{-|x|}/2 \, dx}{\int_{x_{i-1}}^{x_i} e^{-|x|}/2 \, dx} 
			\\[.07in]\notag{}
			&= \frac{(x_{i-1} + 1)e^{-x_{i-1}} - (x_i + 1)e^{-x_i}}{e^{-x_{i-1}} - e^{-x_i}}
		\end{align}
		
		We can use our iterative algorithm to calculate quantization boundaries and intervals and save them for later use. The Conditional Mean Criterion step of the algorithm will use \eqref{eq:laplace_opt_ppoints}.

		\subsubsection{Gaussian Distribution}
		The Gaussian distribution, often referred to as the normal distribution, is ubiquitous in statistical analysis due to its natural occurrence in many physical, biological, and social phenomena. In the context of neural networks, the Gaussian distribution is frequently observed in the distribution of weights, especially in layers that have been subject to regularization techniques.
		
		The Gaussian distribution is characterized by its mean $\mu$ and variance $\sigma^2$, which describe the center and the spread of the data points, respectively. The probability density function (PDF) of the Gaussian distribution is given by:
		\begin{equation}
			f(x|\mu, \sigma^2) = \frac{1}{\sqrt{2\pi\sigma^2}} \exp\left(-\frac{(x - \mu)^2}{2\sigma^2}\right)
		\end{equation} 
		For quantization, these characteristics imply that most data points are concentrated around the mean, with the frequency of occurrence decreasing symmetrically as one moves away from the mean. This property is advantageous for quantization as it allows for finer quantization intervals near the mean where the data points are dense and coarser intervals in the tails where they are sparse.
		
		We aim to minimize the expected quantization error to derive the analytical solution for quantization under the Gaussian distribution. The optimization problem can be formulated as:
		\begin{equation}
			E[Q_E] = \int_{-\infty}^{\infty} (x - Q(x))^2 f(x|\mu, \sigma^2) dx
		\end{equation}  
		where \( Q(x) \) is the quantization function. By exploiting the properties of the Gaussian distribution, we can calculate the optimal quantization levels that minimize the expected error. The resulting quantization scheme will have non-uniformly spaced intervals, with a higher density of intervals near the mean and fewer intervals as the distance from the mean increases, thus accommodating the bell-shaped nature of the Gaussian distribution. The quantization level $y_i$ is computed as follows:
		
		\begin{equation}\label{eq:gauss_opt_ppoints}
			y_i = \sqrt{\frac{2}{\pi}} \cdot \frac{e^{-x_{i-1}^2/2} - e^{-x_i^2/2}}{\erf(x_i/\sqrt{2}) - erf(x_{i-1}/\sqrt{2})},
		\end{equation}

	\noindent	where $\erf$ is the commonly used mathematical error function. 
		
		\begin{equation}
			\text{erf}(x) = \frac{2}{\sqrt{\pi}} \int_{0}^{x} e^{-t^2} dt
		\end{equation}
		
		We can use our iterative algorithm to calculate quantization boundaries and intervals and save them for later use. The Conditional Mean Criterion step of the algorithm will use \eqref{eq:gauss_opt_ppoints}.

		\subsection{Dealing with Non Laplace or Gaussian Distributions}

\begin{figure*}[h]
\centering
\includegraphics[width=\textwidth]{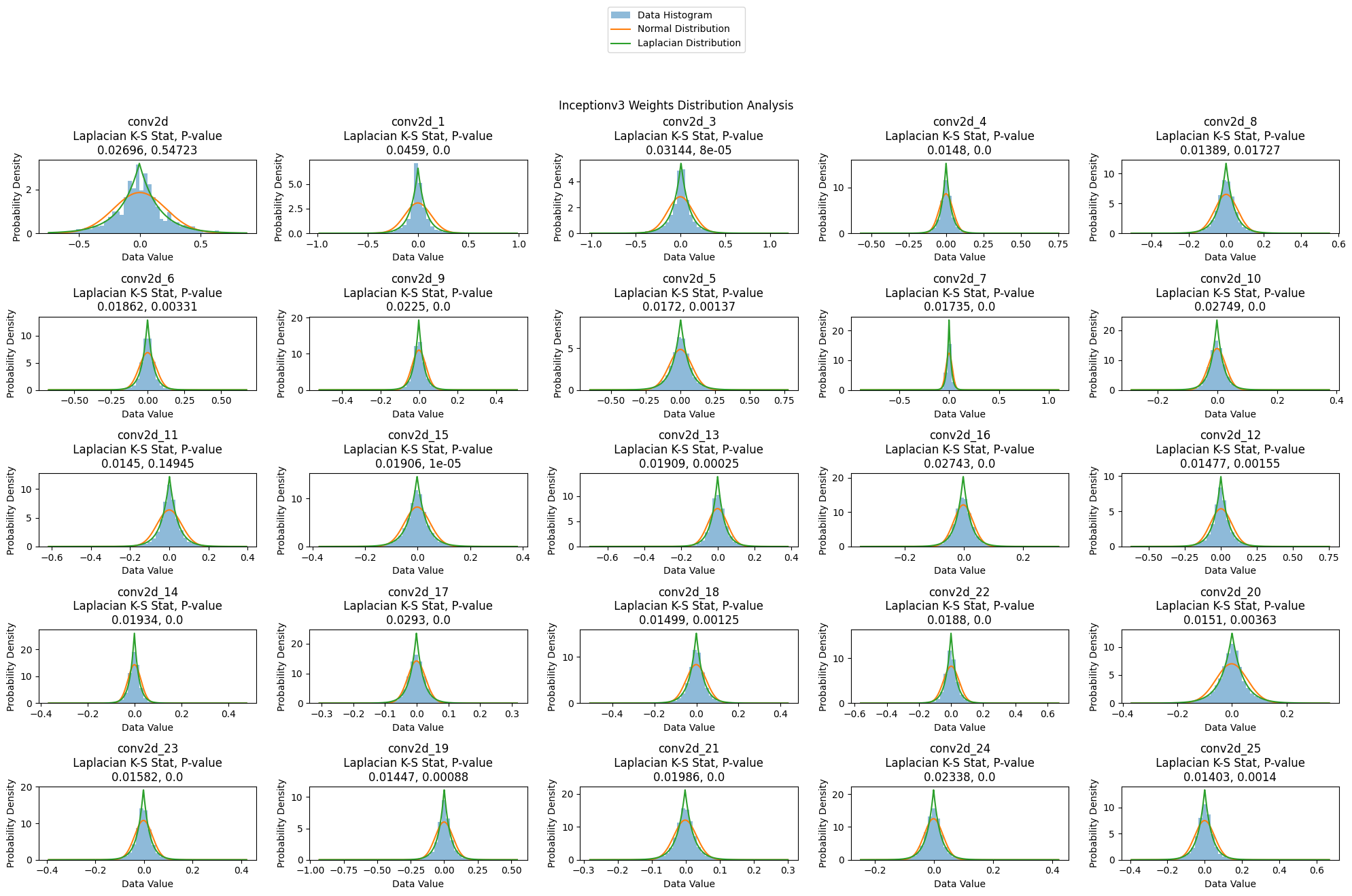}
\caption{Weight distributions in Inceptionv3}
\label{fig_dist}
\end{figure*}

	Weights in most models typically follow either a Laplacian or Normal distribution. Baskin et al. \cite{baskin2021uniq} used Shapiro-Wilk statistics to demonstrate the similarity of ResNet-18 weights to a Normal distribution. We reproduce their results and additionally compare the weights to a Laplacian distribution, as shown in Figure \ref{fig_dist}.

To compare these distributions, we use the Kolmogorov-Smirnov (K-S) test, which measures the maximum difference between the cumulative distribution functions (CDFs) of the actual data and the fitted distributions. We first fit both normal and Laplacian distributions to the data using maximum likelihood estimation to obtain the distribution parameters. A lower K-S statistic indicates a better fit, allowing us to determine whether the weights and activations align more closely with a Gaussian or Laplacian distribution.

Based on the best-fit distribution, we adjust the quantization boundaries and intervals using scale and offset before applying quantization. This process is analogous to the bias correction technique used in previous works \cite{banner2019post}.
		
		\subsection{Quantization for Activations}
		
		Quantizing activations is more technical than quantizing weights. This is because activations depend on input data. Due to computational limitations, only a subset of the data is studied. Recent work has shown the importance of choosing a representative sample \cite{zhang2023selectq}, \cite{williams2023does}, \cite{chen2022climbq}.
		
		Knowledge about the activation function is also critical. For example, the ReLU function (common in CNNs) only outputs positive numbers. This also means the data may no longer be symmetrical. The Fused-ReLU technnique ~\cite{banner2019post} accounts for this by allocating all the bits to the positive side. In this paper, we only quantize the weights. 

		\section{Experiments} \label{sec:experiments}
		In this section, we provide details of the datasets and experimental setup, followed by a comprehensive evaluation of our approach and its comparison with the related methods. 
		
		We present an algorithm that iteratively adjusts the quantization intervals and levels based on the conditional mean criterion and the midpoint criterion as given in  \eqref{eq:laplace_opt_ppoints} and \eqref{eq:gauss_opt_ppoints}. These intervals and levels can all be calculated using the standard parameters. They can later be transformed according to the data's actual mean and standard deviation. 
		
		
		\subsection{Datasets Description}
		
		We compare each method on the popular datasets: ImageNet (ILSVRC12) \cite{deng2009imagenet}, CIFAR10 and CIFAR100 \cite{krizhevsky2009learning}.

		The CIFAR-10 dataset consists of 60000 $32\times 32$ colour images in 10 classes, with 6000 images per class. There are 50000 training images and 10000 test images.
		CIFAR-100 is just like the CIFAR-10, except it has 100 classes containing 600 images each. 
		
		The ImageNet dataset offers 1,000 categories for classification, with 1.2 million training images and 150,000 validation images.
		
		
		For experiments, we perform quantization on ResNet-50,  ResNet-18 and ResNet-20 \cite{He_2016_CVPR} which are pretrained on ImageNet.




		All of the pre-trained model implementations and weights were provided by the cv2 library from Pytorch.

		\subsection{Comparison with other quantization methods}
		
        In our research, we focus on isolating and evaluating individual weight quantization methods to analyze their direct impact on model performance and Mean Squared Error (MSE). 
        
        We use the original FP32 model as our baseline for comparison. For ResNet18, ResNet20, and ResNet50 on CIFAR-10 and CIFAR-100, the baseline models are fine-tuned for 20 epochs prior to quantization. For ResNet18, we also evaluate the effect of directly quantizing the pretrained model without additional fine-tuning.
		
		We compare our method to several closely-related methods: Uniform Quantization like ACIQ \cite{banner2019post} and Non-uniform Quantization like APoT \cite{li2019additive}.
		
		
		For fairness, all the models are compiled in the same way with an adam optimizer and a sparse categorical crossentropy loss. We vary the bitwidths from 4 to 8 bits.

Our method relies on knowing the underlying weight distribution. However, the distribution is not taken as input from the user (unlike competing models, e.g., ACIQ); rather, the knowledge of this distribution is acquired by identifying the best fitting distribution to the data. Also, we feel that this aspect of our work may not be as restrictive as it initially seems, as several prior works (e.g., \cite{baskin2021uniq}, \cite{han2016deep}, and \cite{anderson2018high}) have asserted that the weights in most popular DL architectures indeed have a bell-shaped distribution and the most prominent among them are the Gaussian and Laplacian distributions.

		\subsection{Evaluation Metrics}

		
		
		
		
		\subsubsection{Accuracy}
		
		The models, pretrained on the ImageNet dataset are evaluated on the same task. The model weights are updated and their Top-1 validation accuracy is reported. 
		
		One problem with this measure is that Model Accuracy might improve after quantization due to the introduction of regularization effects, where the reduced precision helps prevent overfitting by smoothing out minor fluctuations in the data. This can sometimes lead to better generalization on unseen data. However, accuracy alone is not the best measure of the effectiveness of quantization. 
		
		\subsubsection{Mean-Squared Error}
		
		In the following figure, we show the Mean Squared Error (MSE) of various quantization methods (Our method, ACIQ, APot) as we go deeper into ResNet50. It can be observed that at the majority of layers, our method has a lower MSE. The results were similar across models like MobileNet-V2 \cite{sandler2018mobilenetv2}, VGG16 \cite{simonyan2014very}, Inceptionv3 \cite{szegedy2016rethinking}.
		
		\begin{figure}[h]
		\centering
    \includegraphics[width=.49\textwidth,page=15]{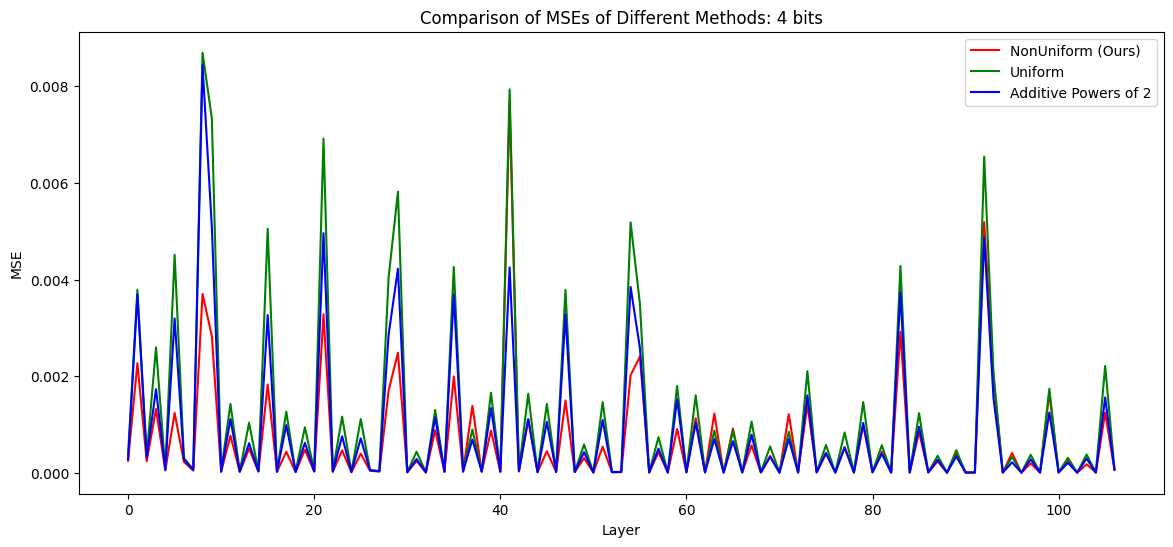}
		\end{figure}
		
		\subsubsection{Results on ImageNet Mini}
		
		The comparison results based on ImageNet are provided in Table \ref{tbl_IMAGENET}

\begin{table}[h!]
    \centering
    \resizebox{.95\linewidth}{!}{
    \begin{tabular}{l l l c c c }
        \toprule
        \textbf{DNN Model} & \makecell{\textbf{FP32} \\ \textbf{Top-1 Acc}} & \textbf{Precision} & \multicolumn{3}{c}{\bf Top 1 Accuracy on ImageNet} \\ \midrule
        & & & \textbf{Ours} & \textbf{Uniform} & \textbf{APoT} \\
        \cmidrule{3-6}
        \multirow{5}{1.2cm}{ResNet18} & \multirow{5}{1.2cm}{69.97} & 4 bits & 46.16 & 44.18 & 47.82 \\ \cmidrule{3-6}
        & & 5 bits & 55.88 & 52.03 & 52.13 \\ \cmidrule{3-6}
        & & 6 bits & 59.14 & 52.49 & 51.98 \\ \cmidrule{3-6}
        & & 7 bits & 61.41 & 52.64 & 52.66 \\ \cmidrule{3-6}
        & & 8 bits & 62.53 & 52.41 & 52.83 \\ \cmidrule{3-6}
        \bottomrule
    \end{tabular}
    }
     \vspace{0.05in}
    \caption{ImageNet Results}
    \label{tbl_IMAGENET}
\end{table}

		\subsubsection{Results on CIFAR10 and CIFAR100}
		
		
		Our method almost always performed better than other methods for the CIFAR-100 task. For CIFAR-10, the results were the same except for the 4-bit variations.
		
		The comparison results based on CIFAR10 and CIFAR100 are provided in  Table \ref{tbl_CIFAR10} and Table \ref{tbl_CIFAR100}, respectively.
		
\begin{table}[h!]
    \centering
    \resizebox{.95\linewidth}{!}{
    \begin{tabular}{l l l c c c }
        \toprule
        \textbf{DNN Model} & \makecell{\textbf{FP32}\\ \textbf{Top-1 Acc}} & \textbf{Precision} & \multicolumn{3}{c}{\bf Top 1 Accuracy on CIFAR-10} \\ \midrule
        & & & \textbf{Ours} & \textbf{Uniform} & \textbf{APoT} \\
        \cmidrule{3-6}
        \multirow{5}{1cm}{ResNet50} & \multirow{5}{1cm}{84.12} & 4 bits & 64.84 & 66.14 & 68.28 \\ \cmidrule{3-6}
        & & 5 bits & 69.97 & 67.04 & 71.06 \\ \cmidrule{3-6}
        & & 6 bits & 66.87 & 66.91 & 69.95 \\ \cmidrule{3-6}
        & & 7 bits & 70.51 & 67.51 & 68.48 \\ \cmidrule{3-6}
        & & 8 bits & 72.18 & 67.62 & 69.31 \\ \midrule
        \multirow{5}{1cm}{ResNet18} & \multirow{5}{1cm}{94.54} & 4 bits & 85.64 & 86.97 & 86.72 \\ \cmidrule{3-6}
        & & 5 bits & 87.69 & 87.42 & 86.81 \\ \cmidrule{3-6}
        & & 6 bits & 87.64 & 87.28 & 87.30 \\ \cmidrule{3-6}
        & & 7 bits & 88.04 & 87.16 & 87.05 \\ \cmidrule{3-6}
        & & 8 bits & 87.82 & 87.24 & 87.34 \\ \midrule
        \multirow{5}{1cm}{ResNet20} & \multirow{5}{1cm}{82.44} & 4 bits & 79.42 & 79.87 & 80.20 \\ \cmidrule{3-6}
        & & 5 bits & 81.96 & 80.96 & 81.31 \\ \cmidrule{3-6}
        & & 6 bits & 82.43 & 82.35 & 82.16 \\ \cmidrule{3-6}
        & & 7 bits & 82.53 & 82.21 & 82.32 \\ \cmidrule{3-6}
        & & 8 bits & 82.46 & 82.38 & 82.48 \\ \cmidrule{3-6}
        \bottomrule
    \end{tabular}
    }
    \vspace{0.05in}
    \caption{CIFAR-10 Results}
    \label{tbl_CIFAR10}
\end{table}

\begin{table}[h!]
    \centering
    \resizebox{.95\linewidth}{!}{
    \begin{tabular}{l l l c c c }
        \toprule
        \textbf{DNN Model} & \makecell{\textbf{FP32}\\ \textbf{Top-1}\\ \textbf{Acc}} & \textbf{Precision} & \multicolumn{3}{c}{\bf Top 1 Accuracy on CIFAR-100} \\ \midrule
        & & & \textbf{Ours} & \textbf{Uniform} & \textbf{APoT} \\
        \cmidrule{3-6}
        \multirow{5}{1.2cm}{ResNet50} & \multirow{5}{1.2cm}{64.07} & 4 bits & 27.45 & 30.51 & 30.77 \\ \cmidrule{3-6}
        & & 5 bits & 43.49 & 28.76 & 29.34 \\ \cmidrule{3-6}
        & & 6 bits & 45.02 & 29.74 & 30.01 \\ \cmidrule{3-6}
        & & 7 bits & 46.62 & 29.22 & 29.46 \\ \cmidrule{3-6}
        & & 8 bits & 46.40 & 29.33 & 29.72 \\ \midrule
        \multirow{5}{1.2cm}{ResNet18} & \multirow{5}{1.2cm}{65.78} & 4 bits & 61.55 & 61.92 & 62.89 \\ \cmidrule{3-6}
        & & 5 bits & 63.43 & 63.04 & 63.36 \\ \cmidrule{3-6}
        & & 6 bits & 63.53 & 63.75 & 63.66 \\ \cmidrule{3-6}
        & & 7 bits & 63.66 & 63.61 & 63.75 \\ \cmidrule{3-6}
        & & 8 bits & 63.45 & 63.55 & 63.93 \\ \midrule
        \multirow{5}{1.2cm}{ResNet20} & \multirow{5}{1.2cm}{51.28} & 4 bits & 47.99 & 43.29 & 41.16 \\ \cmidrule{3-6}
        & & 5 bits & 50.19 & 48.67 & 48.46 \\ \cmidrule{3-6}
        & & 6 bits & 50.52 & 50.44 & 49.88 \\ \cmidrule{3-6}
        & & 7 bits & 51.33 & 50.94 & 50.82 \\ \cmidrule{3-6}
        & & 8 bits & 51.12 & 51.40 & 50.74 \\ \cmidrule{3-6}
        \bottomrule
    \end{tabular}
    }
    \vspace{0.05in}
    \caption{CIFAR-100 Results}
    \label{tbl_CIFAR100}
\end{table}

		\section{Conclusion} \label{sec:conclusion}

		Optimal model compression is one of the focus areas in deep learning to optimize the deployment of complex models. In this paper, we have introduced a novel post-training quantization approach that reduces the computational complexity and storage requirement of deep neural networks (DNNs) without compromising their performance. Our proposed method uses an analytical solution tailored to handle the bell-shaped distribution typically observed in model weights and activations. By determining optimal clipping ranges, quantization intervals, and levels, our approach ensures minimal quantization error, thereby preserving model accuracy.
		
		Our empirical evaluations demonstrate that the proposed quantization scheme effectively reduces model size and computational requirements while maintaining performance, making it particularly suitable for deployment in resource-constrained environments such as mobile devices and embedded systems. 
		
	Future work could focus on optimizing the quantization process through adaptive techniques and integrating our method with other compression strategies to improve performance. We will also assess the impact of our quantization scheme on various models and tasks to validate its effectiveness. Additionally, future directions include calibrating activation quantization using representative training samples and varying bit allocation across layers based on their Mean Squared Error (MSE), assigning more bits to layers with higher MSE. These extensions remain as potential avenues for further research.

\bibliographystyle{IEEEtran}
\bibliography{IEEEexample}

\end{document}